\documentclass[review]{elsarticle}

\usepackage{amssymb}
\usepackage{graphics}
\usepackage{amsmath}
\usepackage{amscd}
\usepackage{amsthm}
\usepackage{mathrsfs}
\usepackage{amsfonts}
\usepackage{amscd}
\usepackage{amsthm}
\usepackage{lscape}
\usepackage{hyperref}
\usepackage{float}
\usepackage{lineno,hyperref}
\usepackage{float}
\usepackage[colorinlistoftodos]{todonotes}
\usepackage[linesnumbered,boxed,lined,ruled,vlined]{algorithm2e}
\modulolinenumbers[5]

\newtheorem{theorem}{Theorem}[section]

\newtheorem{example}{Example}[section]
\newtheorem{definition}{Definition}[section]
\newtheorem{remark}{Remark}[section]
\newtheorem{corollary}{Corollary}[section]

\def\qed{\hbox to 0pt{}\hfill$\rlap{$\sqcap$}\sqcup$}

\journal{Journal of \LaTeX\ Templates}

\begin{document}

\begin{frontmatter}

\title{A New Algorithmic Decision for Categorical Syllogisms via Carroll's Diagrams}

\author[my2address]{Necla Kircali Gursoy}
\ead{necla.kircali.gursoy@ege.edu.tr}

\author[mymainaddress]{Ibrahim Senturk}
\ead{ibrahim.senturk@ege.edu.tr}

\author[mymainaddress]{Tahsin Oner\corref{mycorrespondingauthor}}
\cortext[mycorrespondingauthor]{Corresponding author}
\ead{tahsin.oner@ege.edu.tr}

\author[mymainaddress]{Arif Gursoy}
\ead{arif.gursoy@ege.edu.tr}

\address[my2address]{Tire Kutsan Vocational School, Ege University, Izmir, 35900, Turkey}
\address[mymainaddress]{Department of Mathematics, Ege University, Izmir, 35100, Turkey}

\begin{abstract}
In this paper, we deal with a calculus system SLCD (Syllogistic Logic with Carroll Diagrams), which gives a formal approach to logical reasoning with diagrams, for representations of the fundamental Aristotelian categorical propositions and show that they are closed under the syllogistic criterion of inference which is the deletion of middle term. Therefore, it is implemented to let the formalism comprise synchronically bilateral and trilateral diagrammatical appearance and a naive algorithmic nature. And also, there is no need specific knowledge or exclusive ability to understand as well as to use it. Consequently, we give an effective algorithm used to determine whether a syllogistic reasoning valid or not by using SLCD.  
\end{abstract}


\end{frontmatter}


\section{Introduction}  
\noindent The first idea on syllogisms was produced in the field of proper thinking by the Greek philosopher Aristotle. His idea mainly said that in his Prior Analytics:  \textquotedblleft syllogism is discourse in which, certain things being stated, something other than what is stated follows of necessity from their being so. I mean by the last phrase that they produce the consequence, and by this, that no further term is required from without in order to make the consequence necessary\textquotedblright \cite{jon}.   

A syllogism is a formal logical pattern to obtain a conclusion from the set of premises. A categorical syllogism can be defined as a logical consequence which made up three categorical propositions. It consists of three propositions which are said to be statements or sentences called major premise, minor premise and conclusion, respectively. Each of them has a quantified
relationship between two objects. The position of objects in premises generate a classification called as syllogistic figures. So, there are $4$ different types figures. And, the combination of quantifiers ordering deduces $64$ different combinations in each figure. Therefore,  a categorical syllogistic system consists of $256$ syllogistic moods, $15$ of which are unconditionally and $9$ of them are conditionally; in total $24$ of them are valid. Those syllogisms in the conditional group are also said to be \textit{strengthened}, or valid under \textit{existential import}, which is an explicit assumption of existence of some \textit{S}, \textit{M} or \textit{P}. Then we add a rule to SLCD: \textit{Some X is X when $X$ exists} and consequently, we obtain the formal system SLCD$^\dagger$.

Throughout the centuries, categorical syllogistic system was a paramount part of logic. Innovations in the scope of mathematical logic in the 19th and the beginning of 20th centuries, the situation is changed. However, when  J. \L{}ukasiewicz introduced syllogistic as an axiomatic system built on classical propositional calculus \cite{Lukasiewicz}, the situation became reversed once again. Thereby, categorical syllogistic system plays an important role in the mainstream of contemporary formal logic. Furthermore, \L{}ukasiewicz axiomatization on syllogisms is still open and new ideas rise from time to time. In recent years, the using of syllogisms is studied extensively and investigated under different treatments such as computer science \cite{Kumova, Hartmann}; engineering \cite{Kulik, jet}; artificial intelligence \cite{kryvyi}, \cite{zadeh}; etc. And also, computer science oriented logicians began to take part in \cite{Rocha}.

Using of diagrams in formal logic reasoning has created a spate of interest for years by the reason of needing to visualize complex logic problems that are difficult to understand. For example, at the end of the 1800s, Lewis Carroll used an original diagrammatic scheme to visualize categorical syllogisms in his book \cite{Lewis}. On the contrary using venn diagrams, he used literal diagrams to solve categorical syllogistic problems containing 2-terms, 3-terms and so on. Moreover, using of diagrams in computer systems is a significant topic today because it has potential in offering systems which are more clear and flexible to perform. A common problem of various systems nowadays is that they are complicated which are hard to understand and use. So, we need to use diagrams or other graphical representations to develop more effective and efficient problem solving \cite{nakatsu}. On the contrary the applications of diagrammatic resoning in the cognitive sciences seek a solution how to support learners in complex tasks typically with paper-based or more \textquotedblleft static\textquotedblright \ diagrams \cite{Mayer1, Mayer2}, the applications more typically include how to program a computer to carry out these tasks in artificial intelligence \cite{Glaskow}. Besides, there are also some related works with the using diagrams of syllogisms in different areas such as \cite{moktefi2013beyond, castro2017re, alternativetovennn, manzano}. 

In this paper, we show how the categorical syllogistic statements are expressed using this Carroll literal diagrams. Then, we give a new algorithm deciding whether the syllogism and a strengthened syllogism are valid or not by the help of a calculus system SLCD and SLCD$^{\dagger}$, respectively.

\section{Preliminaries}
In this section, we sketch out notations and terminology which are used during this manuscript.

A categorical syllogism can be defined as a deductive argument consisting of two logical propositions and  a conclusion obtained from these propositions. They contain exactly three terms, each of which occurs in exactly two of the constituent propositions and the conclusion, where the propositions and the conclusion each has a quantified relationship between two terms. The objects in a categorical proposition are related with following four distinct forms as in Table \ref{tab-1}.

\begin{table}[h!]
                    \centering 
                    \caption{Categorical Syllogistic Propositions} 
                    \label{tab-1} 
                    \begin{tabular}{c c c }
                        \hline
                        Symbol & Statements & Generic Term  \\
                        \hline
                        $A$ & All $X$ are $Y$	& Universal Affirmative\\
                        $E$ & No $X$ are $Y$	& Universal Negative\\
                        $I$ & Some $X$ are $Y$	& Particular Affirmative\\
                        $O$ & Some $X$ are not $Y$	& Particular Negative\\
                        \hline
\end{tabular}
\end{table}

For any syllogism, the categorical propositions are composed of three terms, a subject term, a predicate term, and a middle term: the subject term is the subject of the conclusion and denoted by $S$; the predicate term modifies the subject in the conclusion and denoted by $P$, and the middle term which occurs in the two premises and links the subject and predicate terms and noted by $M$. The subject and predicate terms occur in different premises but the middle term occurs once in each premise.  The premise which consists of the predicate term and the middle term is called the \textit{major premise}; the premise which consists of subject term and the middle term is called the \textit{minor premise}.

Categorical syllogisms are grouped into $4$ different ways, which are traditionally called figures,  depending on the position of the term-variables $S$, $P$ and $M$ in Table \ref{tab-2}.

\begin{table}[h!]
                    \centering 
                    \caption{Categorical Syllogistic Figures} 
                    \label{tab-2} 
\begin{tabular}
[c]{|c|c|c|c|}\hline
Major & Minor & Conclusion & Figure\\\hline\hline
$M-P$ & $S-M$ & $S-P$ & 1\\\hline
$P-M$ & $S-M$ & $S-P$ & 2\\\hline
$M-P$ & $M-S$ & $S-P$ & 3\\\hline
$P-M$ & $M-S$ & $S-P$ & 4\\\hline
\end{tabular}
\end{table}

Aristotle identified only  the first three figures, but the last one was discovered in the middle ages. He searched each mood and figure in terms of whether it was valid or not. After, he obtained some common properties of these syllogisms, which are called rules of deduction. These rules are as follows:

$\textbf{Step 1:}$ Relating to premises irrespective of conclusion or figure:
\begin{itemize}
\item[(a)]No inference can be made from two particular premises.
\item[(b)]No inference can be made from two negative premises.
\end{itemize}

$\textbf{Step 2:}$ Relating to propositions irrespective of figure:

\begin{itemize}
\item[(a)]If one premise is particular, the conclusion must be particular.
\item[(b)]If one premise is negative, the conclusion must be negative.
\end{itemize}

$\textbf{Step 3:}$ Relating to distribution of terms:

\begin{itemize}
\item[(a)]The middle term must be distributed at least once.
\item[(b)]A predicate distributed in the conclusion must be distributed in the major premise.
\item[(c)]A subject distributed in the conclusion must be distributed in the minor premise.
\end{itemize}

In categorical syllogistic system, there are 64 different syllogistic forms for each figure. These are called \textit{moods}. Therefore, the categorical syllogistic system is composed of 256 possible syllogisms. Only 24 of them are valid in this system. And  they divided into two groups of 15 and of 9.\\

The syllogisms in the first group are valid \textit{unconditionally} which are given in Table \ref{tab-3}.

\begin{table}[h!]
               \centering 
               \caption{Unconditionally Valid Forms} 
               \label{tab-3} 
               \begin{tabular}{ c c c c}
                   \hline
                   Figure I & Figure II & Figure III & Figure IV \\
                   \hline
                   $AAA$ & $EAE$ & $IAI$ & $AEE$\\
                   $EAE$ & $AEE$ & $AII$ & $IAI$\\
                   $AII$ & $EIO$ & $OAO$ & $EIO$\\
                   $EIO$ & $AOO$ & $EIO$ &    \\
                   \hline
 \end{tabular}
\end{table}
The syllogisms in the second group called \textit{strengthened syllogism} are valid \textit{conditionally} or valid \textit{existential import} ,which is an explicit supposition of being of some terms, are shown in Table \ref{tab-4}.

\pagebreak

\begin{table}[h!]
                   \centering 
                   \caption{Conditionally Valid Forms} 
                   \label{tab-4} 
                   \begin{tabular}{c c c c c}
                       \hline
                       Figure I & Figure II & Figure III & Figure IV & Necessary Condition\\
                       \hline
                       $AAI$ & $AAO$ &     & $AEO$ & \textit{S} exists\\
                     $EAO$	& $EAO$ &     &     & \textit{S} exists\\
                       	&     & $AAI$ & $EAO$ & \textit{M} exists\\
                           &     & $EIO$ &     & \textit{M} exists\\
                           &     &     & $AAI$ & \textit{P} exists\\
                       \hline
\end{tabular}
\end{table}

\section{Representation of Categorical Syllogisms via Carroll's Diagrams and a Calculus System SLCD}

Carroll's diagrams, thought up in 1884, are Venn-type diagrams where the universes are represented with a square. Nevertheless, it is not clear whether Carroll studied his diagrams independently or as a modification of John Venn's. Carroll's scheme looks like a productive method summing up several developments that have been introduced by researchers studying in this area. 

For categorical syllogistic system, we describe an homomorphic mapping between the categorical syllogistic propositions and the Carroll's diagrams. Let $X$ and $Y$ be two terms and let $X'$ and $Y'$ be complements of $X$ and $Y$, respectively. For two-terms, Carroll divides the square into four cells, by this means he gets the so-called bilateral diagram, as shown in Table \ref{tab-5}.

\begin{table}[h!]
\centering
\caption{Relation of Two Terms}
\label{tab-5}
\begin{tabular}{|c|c|c|}
\hline
& $X'$ & $X$  \\
\hline
$Y'$ & $X'Y'$ & $XY'$   \\
\hline
$Y$ & $X'Y$ & $XY$    \\
\hline
\end{tabular}
\end{table}

Each of these four cells can have three possibilities, when we explain the relations between two terms. They can be $0$ or $1$ or \textit{blank}.
In this method, $0$ means that there is no element intersection cell of two elements, $1$ means that it is not empty and \textit{blank} cell means that we don't have any idea about the content of the cell, it could be $0$ or $1$.

\begin{figure}[h!]
\centering {\scalebox{0.70}{
\includegraphics[ ]{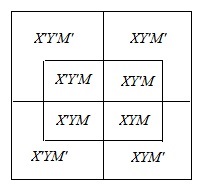}}}%
\caption{Relations of three terms}
\label{fig-1}
\end{figure}

As above method, let $X$, $Y$, and $M$ be three terms and $X'$, $Y'$, and $M'$ be complements of $X$, $Y$, and $M$, respectively. For examining all relations between three terms, he added one more square in the middle of bilateral diagram which is called the trilateral diagram, as in Figure \ref{fig-1}.

Each cell in a trilateral diagram is marked with a $0$, if there is no element and is marked with a $\textbf{I}$ if it is not empty and another using of $\textbf{I}$, it could be on the line where the two cell is intersection, this means that at least one of these cells is not empty. So, $\textbf{I}$ is different from $1$. In addition to these, if any cell is \textbf{blank}, it has two possibilities, $0$ or $\textbf{I}$.

To obtain the conclusion of a syllogism, the knowledge of two premises are carried out on a trilateral diagram. This presentation is more useful for the elimination method than the Venn diagram view. In this way, one can observe the conclusion of the premises truer and quicker from a trilateral diagram. By dint of this method, we demean the data from a trilateral diagram to a bilateral diagram, involving only two terms that should occur in the conclusion and consequently eliminating the middle term.

This method can be used in accordance with the rules below \cite{Lewis}:

\noindent\textit{\textbf{First Rule:}} $0$ and $\textbf{I}$ are fixed up on trilateral diagrams.

\noindent\textit{\textbf{Second Rule:}} If the quarter of the trilateral diagram contains a \textquotedblleft$\textbf{I}$\textquotedblright \ in either cell, then it is certainly occupied, and one may mark the corresponding quarter of the bilateral diagram with a \textquotedblleft$1$\textquotedblright \ to indicate that it is occupied.

\noindent\textit{\textbf{Third Rule:}} If the quarter of the trilateral diagram contains two \textquotedblleft$0$\textquotedblright s, one in each cell, then it is certainly empty, and one may mark the corresponding quarter of the bilateral diagram with a \textquotedblleft$0$\textquotedblright \ to indicate that it is empty. 

We obtain the required conclusion of a syllogism by using of these rules. The effect of Carroll's method of transfer, unknown to Venn, could not be underestimated. It only shows how to extract the conclusion from the premises of a syllogism \cite{moktefi2012history}.

Now, we give the set theoretical representation of syllogistic
arguments by means of bilateral diagrams. To build such a model, we draw from Carroll's diagrammatic method.  We give a definition of a map which corresponds each bilateral diagram to a set. Eventually, our main purpose is to construct a complete bridge between sets and categorical syllogisms such as Table \ref{tab-6}.

 \begin{table}[h]
                   \centering 
                   \caption{\textit{The Paradigm for the representation of syllogistic arguments by using sets}} 
                   \label{tab-6} 
\begin{tabular}{c|c|c|c}
                       	& LOGIC& DIAGRAMS & SETS  \\
                       \hline
                       PREMISES& Propositions  & $\xrightarrow{Translate}$& Sets\\
                   & & & $\downarrow$\\
                      CONCLUSIONS & Propositions	& $\xleftarrow{Translate}$  & Sets \\

                   \end{tabular}
               \end{table}

Let $X$ and $Y$ be two terms and their complements are denoted by $X'$ and $Y'$, respectively. Assume that $p_i$ shows a possible form of any bilateral diagram, such that $1\leq i \leq k$, where $k$ is the number of possible forms of bilateral diagram, as in Table \ref{tab-7}.  

\begin{table}[h]
	     \centering 
	     \caption{Bilateral diagram for a quantity relation between $X$ and $Y$} 
	     \label{tab-7} 
	     \begin{tabular}{|c|c|c|}
	        \hline
	        $p_i$ & $X'$ & $X$  \\
	        \hline
	        $Y'$   & $n_1$ & $n_2$   \\
	        \hline
	        $Y$ & $n_3$ & $n_4$   \\
	        \hline
	      \end{tabular}
\end{table}
\noindent where $n_1, n_2, n_3, n_4\in\{0,1\}$. During this paper,  $R_{(A)}$,  $R_{(E)}$,  $R_{(I)}$ and  $R_{(O)}$  correspond ``$All$'', \textquotedblleft$No$\textquotedblright, \textquotedblleft$Some$\textquotedblright and \textquotedblleft$Some-not$\textquotedblright statements, respectively.

\begin{example}\label{example1}
We analyze $\textit{``No X are Y''}$ statement means that there is no element in the intersection cell of $X$ and $Y$. We show it in the following bilateral diagram as in Table \ref{tab-8}. From Table \ref{tab-8}, we obtain all possible bilateral diagrams which have $0$ in the intersection cell of $X$ and $Y$. So, Table \ref{tab-9} shows all possible forms of $\textit{``No X are Y"}$.\\
\begin{table}[h!]
                       \centering 
                       \caption{\textit{Bilateral diagram for ``$No$ $X$ $are$ $Y$"}} 
                        \label{tab-8} 
                       $R_{(A)}=$
                       \begin{tabular}{|c|c|c|}                   \hline
                           	& $X'$ & $X$  \\
                           \hline
                           $Y'$   &  &    \\
                           \hline
                           $Y$	&  & 0    \\
                           	\hline
                       \end{tabular}
\end{table}

\begin{table}[h!]
	   \centering
	   \caption{\textit{All possible forms of ``$All$ $X$ $are$ $Y$"}}
	   \label{tab-9}
           \begin{tabular}{|c|c|c|}
              \hline
              $\boldsymbol{p_1}$ & $X'$ & $X$  \\
              \hline
              $Y'$ & 0 & 0   \\
              \hline
              $Y$ & 0 & 0    \\
              \hline
            \end{tabular}\ \
            \begin{tabular}{|c|c|c|}
              \hline
              $\boldsymbol{p_2}$ & $X'$ & $X$  \\
              \hline
              $Y'$ & 0 & 0   \\
              \hline
              $Y$ &1 & 0    \\
              \hline
            \end{tabular}\ \
            \begin{tabular}{|c|c|c|}
              \hline
              $\boldsymbol{p_3}$ & $X'$ & $X$  \\
              \hline
              $Y'$ &  0 & 1   \\
              \hline
              $Y$ & 0 & 0    \\
              \hline
            \end{tabular}\ \
            \begin{tabular}{|c|c|c|}
              \hline
              $\boldsymbol{p_4}$ & $X'$ & $X$  \\
              \hline
              $Y'$ & 1 & 0   \\
              \hline
              $Y$ & 0 & 0    \\
              \hline
            \end{tabular}\ \
            
            \begin{tabular}{|c|c|c|}
              \hline
              $\boldsymbol{p_5}$ & $X'$ & $X$  \\
              \hline
              $Y'$ &  0 & 1   \\
              \hline
              $Y$ & 1 & 0    \\
              \hline
            \end{tabular}\ \
            \begin{tabular}{|c|c|c|}
              \hline
              $\boldsymbol{p_6}$ & $X'$ & $X$  \\
              \hline
              $Y'$ & 1 & 0   \\
              \hline
              $Y$ & 1 & 0    \\
              \hline
            \end{tabular}\ \
            \begin{tabular}{|c|c|c|}
              \hline
              $\boldsymbol{p_7}$ & $X'$ & $X$  \\
              \hline
              $Y'$ & 1 & 1   \\
              \hline
              $Y$ & 0 & 0    \\
              \hline
            \end{tabular}\ \
            \begin{tabular}{|c|c|c|}
              \hline
              $\boldsymbol{p_8}$ & $X'$ & $X$  \\
              \hline
              $Y'$ & 1 & 1   \\
              \hline
              $Y$ & 1 & 0    \\
              \hline
            \end{tabular}
\end{table}

\end{example}

Now in order to define a relation between bilateral diagrams and sets, let us form a set consisting of numbers which correspond to possible forms that each bilateral diagram possesses. For this aim, we firstly define a value mapping in which each possible bilateral diagram corresponds to exactly one value.

\begin{definition}\label{definition1}\cite{rus} Let $p_i$ be a possible bilateral diagram  and $n_i$ be the value that the $i$-th cell possesses. The $r^{\mathit{val}}_j$ corresponds to the value of $p_i$ which is calculated by using the formula
$$r^{\mathit{val}}_j=\sum_{i=1}^4 2^{(4-i)}n_i, \  \  \ 1\leq j\leq k,$$
where $k$ is the number of all possible forms. 
\end{definition}

\begin{definition}
Let $R^{\mathit{set}}$ be the set of the values which correspond to all possible forms of any bilateral diagram; that is $R^{\mathit{set}}=\{r^{\mathit{val}}_j: 1\leq j \leq k,\text{$k$ is the number}$ $\text{of all possible forms}\}$. The set of all these $R^{\mathit{set}}$'s is denoted by $\mathcal{R}^{\mathit{Set}}$.
\end{definition}

\begin{corollary}
We obtain the set representations of all categorical propositions as follows:
\begin{itemize}
\item[-] \textit{All X are Y:} It means that $X$ intersection with $Y'$ cell is empty set. We can illustrate this statement as Table \ref{tab-10}.
\begin{table}[h!]
\centering
\caption{\textit{$X$ intersection with $Y'$ is empty set}}
\label{tab-10}
$R_{(A)}=$\begin{tabular}{|c|c|c|}
\hline
& $X'$ & $X$  \\
\hline
$Y'$   &  &0    \\
\hline
$Y$	&  &     \\
\hline
\end{tabular}
\end{table}

\noindent From Table \ref{tab-10}, we obtain all possible forms as the same method in Example \ref{example1}. By the help of Definition \ref{definition1}, the set representation of ``\textit{All X are Y}" corresponds to the $R^{\mathit{set}}_{(A)}=\{0,1,2,3,8,9,10,11\}$.

\item[-]\textit{No X are Y:}There is no element in the intersection cell of $X$ and $Y$ as  Table \ref{tab-11}.
\begin{table}[h]
\centering
\caption{\textit{$X$ intersection with $Y$ is empty set}}
\label{tab-11}
$R_{(E)}=$\begin{tabular}{|c|c|c|}
\hline
& $X'$ & $X$  \\
\hline
$Y'$   &  &   \\
\hline
$Y$	&  &0     \\
\hline
\end{tabular}
\end{table}

\noindent By Example \ref{example1}, we have all possible forms of ``\textit{No X are Y}". Then, we obtain
$R^{\mathit{set}}_{(E)}=\{0,2,4,6,8,10,12,14\}$.
\newpage
\item[-]\textit{Some X are Y:} There is at least one element in the intersection $X$ and $Y$ as Table \ref{tab-12}.
\begin{table}[h!]
\centering
\caption{\textit{$X$ intersection $Y$ has at least one element}}
\label{tab-12}
$R_{(I)}=$\begin{tabular}{|c|c|c|}
\hline
& $X'$ & $X$  \\
\hline
$Y'$   &  &    \\
\hline
$Y$	&  & 1     \\
\hline
\end{tabular}
\end{table}

By using the possible bilateral diagrams of $R_{(I)}$, we have $R^{\mathit{set}}_{(I)}=\{1,3,5,7,9,11,13,15\}$.

\item[-]\textit{Some X are not Y:}
If some element of $X$ are not $Y$, then they have to be in $Y'$. So, the intersection cell of $X$ and $Y'$ is not empty as Table \ref{tab-13}.
\begin{table}[h!]
\centering
\caption{\textit{$X$ intersection $Y'$ has at least one element}}
\label{tab-13}
$R_{(O)}=$\begin{tabular}{|c|c|c|}
\hline
& $X'$ & $X$  \\
\hline
$Y'$   &  & $1$    \\
\hline
$Y$	&  &     \\
\hline
\end{tabular}
\end{table}

\noindent From the bilateral diagram of $R_{(O)}$, we get  $R^{\mathit{set}}_{(O)}=\{4,5,6,7,12,13,14,15\}$.
\end{itemize}
\end{corollary}

Let's consider the relationship between the possible bilateral diagrams of the categorical syllogisms before discussing of the categorical syllogisms via Carroll's diagrams.

\begin{example}
Let $p_i$ and $p_j$ be two possible forms of the bilateral diagrams of major and minor premises, respectively. We take the possible forms of bilateral diagrams as Table \ref{tab-14}.

\begin{table}[h]
                  \centering
                  \caption{The possible forms of bilateral diagrams}
                  \label{tab-14}
                  $p_i=$
                  \begin{tabular}{|c|c|c|}                   \hline
                & $P'$ & $P$  \\
                 \hline
                $M'$   & 1  & 0    \\
                \hline
                $M$	& 0   & 0    \\
                                                  	
                 \hline
                \end{tabular}
                \ \ and \ \ \  $p_j=$
                \begin{tabular}{|c|c|c|}                   \hline
                & $S'$ & $S$  \\
                 \hline
                $M'$   & 0  & 1    \\
                \hline
                $M$	& 0   & 0    \\
                                                  	
                 \hline
                \end{tabular}
\end{table}

We input the data on trilateral diagram as in Figure \ref{fig-2}.

\begin{figure}[h!]
\centering
\caption{The relation of two possible forms}
\label{fig-2}
{\scalebox{0.70}{
\includegraphics[]{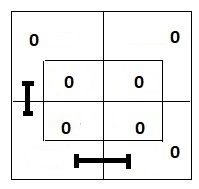}}}%

\end{figure}

By using the elimination method, we obtain the relation between $S$ and $P$ as Table \ref{tab-15}.
\begin{table}[h]
                     \centering
                  	\caption{The relation between $S$ and $P$}
                  	\label{tab-15}
                     $p_k=$
                     \begin{tabular}{|c|c|c|}                   \hline
                     & $P'$ & $P$  \\
                      \hline
                     $S'$   & 0  & 0    \\
                     \hline
                     $S$	& 1   & 0    \\
                                                       	
                      \hline
                     \end{tabular}
\end{table}

$r_i=8$ corresponds to possible form $p_i$, and $r_j=4$ corresponds to possible form $p_j$, then we obtain that $r_k=2$ corresponds to $p_k$ that is a possible conclusion.
\end{example}

Let $r^{\mathit{val}}_i$ and $r^{\mathit{val}}_j$ be the numbers corresponding to possible forms of bilateral diagrams which have a common term. Then we can get the relation between two other terms by using this method.

After these examples, we try to generalize them by a formula. 

\begin{definition}
The syllogistic possible conclusion mapping, denoted $\ast$, is a mapping which gives us the deduction set of possible forms of major and minor premises sets.
\end{definition}

\begin{theorem}
Let $r^{\mathit{val}}_i$ and $r^{\mathit{val}}_j$ correspond to the numbers of possible forms of major and minor premises, respectively. Then, $r^{\mathit{val}}_i\ast r^{\mathit{val}}_j$ equals the value given by the intersection of row and column numbers corresponding to $r^{\mathit{val}}_i$ and $r^{\mathit{val}}_j$ in Table \ref{tab-16}.
\end{theorem}

\begin{table}[h]
              \centering 
              \caption{Operation table} 
              \label{tab-16} 
          {\small
          \begin{tabular}{|c|c|c|c|c|c|c|c|c|c|c|c|c|c|c|c|c|}
          \hline 	
           $\ast$& 0 & 1 & 2 & 3 & 4 & 8 & 12 & 5 & 10 & 6 & 9 & 7 & 11 & 13 & 14 & 15 \\
           \hline
           0& 0 &  &  &  &  &  &  &  &  &  &  &  &  &  &  &   \\
           \hline
           1&  & 1 & 4 & 5 &  &  &  &  &  &  &  &  &  &  &  &  \\
           \hline
           2&  & 2 & 8  & 10 &  &  &  &  &  &  &  &  &  &  &  &  \\
           \hline
           3&  & 3 & 12 & $H$ &  &  &  &  &  &  &  &  &  &  &  &  \\
           \hline
           4&  &  &  &  & 1 & 4 & 5 &  &  &  &  &  &  &  &  &  \\
          \hline 	
           8&  &  &  &  & 2 & 8 & 10 &  &  &  &  &  &  &  &  &  \\
           \hline
           12&  &  &  &  & 3 & 12 & $H$ &  &  &  &  &  &  &  &  &  \\
           \hline
           5&  &  &  &  &  &  &  & 1 & 4 & 5 & 5 & 5 & 5 & 5 & 5 & 5 \\
           \hline
           10&  &  &  &  &  &  &  & 2 & 8 & 10 & 10 & 10 & 10 & 10 & 10 & 10  \\
           \hline
           6&  &  &  &  &  &  &  & 3 & 12 & 9 & 6 & 11 & 14 & 7 & 13 &  15\\
           \hline
           9&  &  &  &  &  &  &  & 3 & 12 & 6 & 9 & 7 & 13 & 11 & 14 & 15 \\
           \hline
           7&  &  &  &  &  &  &  & 3 & 12 & 13 & 7 & $H_4$ & $H'_3$ & 7 & 13 & $H'_1$  \\
           \hline
           11&  &  &  &  &  &  &  & 3 & 12 & 14 & 11 & $H_3$ & $H'_4$ & 11 & 14 & $H'_2$ \\
           \hline
           13&  &  &  &  &  &  &  & 3 & 12 & 7 & 13 & 7 & 13 & $H_4$ & $H'_3$ & $H'_1$ \\
           \hline
           14&  &  &  &  &  &  &  & 3 & 12 & 11 & 14 & 11 & 14 & $H_3$ & $H'_4$ & $H'_2$ \\
           \hline   	   	                      	
           15&  &  &  &  &  &  &  & 3 & 12 & 15 & 15 & $H_1$ & $H_2$ & $H_1$ & $H_2$ & $H$ \\
           \hline
          \end{tabular}}
\end{table}
In the Table \ref{tab-16},  considering possible conclusion operation, some possible forms of premises have more than one possible conclusions, given as below:
\begin{gather*}
           H=\{6, 7, 9, 11, 13, 14, 15\},\ H_1=\{7, 11, 15\},\ H_1'=\{6, 7, 9, 11, 13, 15\},\\
           H_2=\{13, 14, 15\},\ H_2'=\{11, 14, 15\},\ H_3=\{6, 7, 11, 14, 15\},\\
           H_3'=\{6, 7, 13, 14, 15\},\ H_4=\{7, 9, 11, 13, 15\},\ H_4'=\{9, 11, 13, 14, 15\}
\end{gather*}
Therefore, we scrutinise all possible cases between two terms and their conclusions.

Note that, Table \ref{tab-16} is used as $Syllogistic\_Mapping()$ subalgorithm in Section \ref{sec4}.

\begin{definition}
Universes of values sets of major premises, minor premises, and conclusions are denoted by $\mathcal{R}^{\mathit{set}}_{\textit{Maj}}$, $\mathcal{R}^{\mathit{set}}_{\textit{Min}}$ and $\mathcal{R}^{\mathit{set}}_{\textit{Con}}$, respectively.
\end{definition}

Let $R^{\mathit{set}}_{(k)}$ be an element of $\mathcal{R}^{\mathit{set}}_{\textit{Maj}}$ and $R^{\mathit{set}}_{(l)}$ be an element of $\mathcal{R}^{\mathit{set}}_{\textit{Min}}$. The main problem is what the conclusion of these premises is. In syllogistic, we have some patterns which are mentioned in Table \ref{tab-3} and Table \ref{tab-4} above. Now, we explain them by using bilateral diagrams with an algebraic approach.
\begin{definition}
The syllogistic mapping, denoted by $\circledast$, is a mapping which gives us the conclusion of the major and the minor premises as Table \ref{tab-17}.
\begin{table}[h!]
\centering
\caption{The conclusion of the major and the minor premises} 
\label{tab-17} 

 	     \begin{tabular}{|c|c|c|}
 	        \hline
 	        & $P'$ & $P$  \\
 	        \hline
 	        $M'$   &  &   \\
 	        \hline
 	        $M$ & $$ & $$   \\
 	        \hline
 	      \end{tabular}
 	      $\circledast$ \begin{tabular}{|c|c|c|}
 	       	        \hline
 	       	        & $S'$ & $S$  \\
 	       	        \hline
 	       	        $M'$   &  &    \\
 	       	        \hline
 	       	        $M$ & &    \\
 	       	        \hline
 	       	      \end{tabular}
 	       	      = \begin{tabular}{|c|c|c|}
 	       	       	        \hline
 	       	       	        & $P'$ & $P$  \\
 	       	       	        \hline
 	       	       	        $S'$   &  &   \\
 	       	       	        \hline
 	       	       	        $S$ &  &   \\
 	       	       	        \hline
 	       	       	      \end{tabular}
\end{table}
\end{definition}
\begin{theorem}\label{theorem4.12}
Let $R^{\mathit{set}}_{(k)}=\{r^{\mathit{val}}_{k_1},\dots, r^{\mathit{set}}_{k_n}\}$ and $R^{\mathit{set}}_{(l)}=\{r^{\mathit{val}}_{l_1},\dots, r^{\mathit{val}}_{l_t}\}$  two sets corresponding to the Major and the Minor premises. Then $\circledast: \mathcal{R}^{\mathit{set}}_{\textit{Maj}}\times\mathcal{R}^{\mathit{set}}_{\textit{Min}}\rightarrow \mathcal{R}^{\mathit{set}}_{\textit{Con}}$
$$R^{\mathit{set}}_{(k)} \circledast R^{\mathit{set}}_{(l)}:= \bigcup^n_{j=1} \bigcup^t_{i=1} r^{\mathit{val}}_{k_j}\ast r^{\mathit{val}}_{l_i}$$
is the conclusion of the premises $R^{\mathit{set}}_{(k)}$ and $R^{\mathit{set}}_{(l)}$.
\end{theorem}

\begin{theorem}\cite{senturkoner}
A syllogism is valid if and only if it is provable in \textit{SLCD}.
\end{theorem}

\begin{remark}
For conditional valid forms, we need an addition rule which is \textit{\textquotedblleft Some $X$ are $X$"}. We can use above Theorem by taking into consideration this rule.
\end{remark}
\begin{remark}
Let SLCD be noted calculus system. If the rule \textit{\textquotedblleft Some X are X when X exists"} (i.e.,  $\vdash\boldsymbol{I}_{XX}$) is added to SLCD, then the calculus system SLCD is denoted by  $\textit{SLCD}^{\dagger}$.
\end{remark}

\begin{definition}\cite{senturkoner}
Let $R_{(k)}$ be the bilateral diagram presentation of the premise. The \textit{transposition} of a premise is the symmetric positions with respect to the main diagonal. It is shown by $Trans(R_{(k)})$.
\begin{eqnarray*}
Trans:\mathcal{R}^{\mathit{set}}&\rightarrow& \mathcal{R}^{\mathit{set}},\\
{R}^{\mathit{set}}_{(k)}&\rightarrow& Trans({R}^{\mathit{set}}_{(k)}) =\{r^{\mathit{val}}_{k^T_1},\dots, r^{\mathit{set}}_{k^T_n}\}.\nonumber
\end{eqnarray*}
\end{definition}

\newpage
\begin{theorem}\label{theorem4.17}\cite{senturkoner}
Let $R^{\mathit{set}}_{(k)}=\{r^{\mathit{val}}_{k_1},\dots, r^{\mathit{set}}_{k_n}\}$ and $R^{\mathit{set}}_{(l)}=\{r^{\mathit{val}}_{l_1},\dots, r^{\mathit{val}}_{l_t}\}$ be two sets to correspond to the Major and the Minor premises values sets and $R^{\mathit{set}}_{(s)}=\{r^{\mathit{val}}_{s_1},\dots, r^{\mathit{set}}_{s_m}\}$ be set to correspond to the constant set values which means \textquotedblleft Some S are S", \textquotedblleft Some M are M" and \textquotedblleft Some P are P". Then $\circledast^{\dagger}: \mathcal{R}^{\mathit{set}}_{\textit{Maj}}\times\mathcal{R}^{\mathit{set}}_{\textit{Min}}\rightarrow \mathcal{R}^{\mathit{set}}_{\textit{Con}}$
$$
R^{\mathit{set}}_{(k)} \circledast^{\dagger} R^{\mathit{set}}_{(l)}:=
\begin{cases}
\bigcup^n_{j=1} \bigcup^t_{i=1} \bigcup^m_{h=1} (r^{\mathit{val}}_{k_j}\ast (r^{\mathit{var}}_{s_h}\ast r^{\mathit{var}}_{l^T_i})), \; \; & \textit{If S exists}, \\
\bigcup^n_{j=1} \bigcup^t_{i=1} \bigcup^m_{h=1} (r^{\mathit{val}}_{k_j}\ast (r^{\mathit{var}}_{l_i}\ast r^{\mathit{var}}_{s_h} )), \; \;  & \textit{If M exists}, \\
\bigcup^n_{j=1} \bigcup^t_{i=1} \bigcup^m_{h=1} ((r^{\mathit{var}}_{s_h} \ast r^{\mathit{val}}_{k^T_j})\ast r^{\mathit{var}}_{l_i}), \; \;  & \textit{If P exists}.
\end{cases}$$
is the conclusion of the premises $R^{\mathit{set}}_{(k)}$ and $R^{\mathit{set}}_{(l)}$  under the conditions \textit{S exists}, \textit{M exists} or \textit{P exists}.
\end{theorem}

\begin{theorem}\cite{senturkoner}
A strengthened syllogism is valid if and only if it is provable in \textit{SLCD}$^{\dagger}$.
\end{theorem}

\section{An Algorithmic Decision for Categorical Syllogisms in SLCD}\label{sec4}
In this part of the manuscript, we give an algorithm to decide whether a categorical syllogism is valid or not in the calculus system SLCD or SLCD$^{\dagger}$ at the first time in the literature. \\
Global variables used to all functions are below:\\
$Conc[\ ][\ ]:$ two dimensional array, set of all possible bilateral diagrams \\
$Const\_set:$ constant set for each condition S exists, M exists and P exists

$\bullet$ Algorithm Syllogism:\\
This is the main algorithm. In this algorithm, $MPSM()$ and $Decision()$ subalgorithms are run for each state (Unconditional, S exists, M exists and P exists) and for each figure (Figure1, Figure2, Figure3 and Figure4). This algorithm sends related figure as parameter to subalgorithm $MPSM()$ and also, it sends the related state and figure as parameters to subalgorithm $Decision()$.

\begin{algorithm}[H]
\DontPrintSemicolon
\caption{Syllogism\label{A1}}
\KwData{All states for each Figure}
\KwResult{Obtain the Conclusion set of syllogisms and make a decision for syllogisms whether \textquotedblleft Valid" or \textquotedblleft Invalid".}
\BlankLine
\emph{\textbf{Syllogism()}} \;
\ForEach{$cond$ in $Conditions\{Unconditional, S\_exists, M\_exists, P\_exists\}$}{
	\ForEach{$fig$ in $Figures\{Figure1, Figure2, Figure3, Figure4\}$}{
		$MPSM(fig)$\;
		$Decision(cond, fig)$\;
	}
}
\end{algorithm}
\vspace{0.5cm}

$\bullet$ Subalgorithm MPSM:\\
This algorithm determines the positions of the subject term, middle term and predicate term with respect to the figure parameter as input.\\

\begin{algorithm}[H]
\DontPrintSemicolon
\caption{MPSM \label{A2}}
\KwData{The specified figure}
\KwResult{Positions of the major and minor terms are determined}
\BlankLine

\emph{\textbf{MPSM(fig)}} \;

\uIf{$fig= ``Figure 1"$}{
	$mj_1 = ``M"; mj_2 = ``P"$\;
	$mn_1 = ``S"; mn_2 = ``M"$\;}
    
   \uElseIf{$fig=``Figure 2"$}{
    $mj_1 = ``P"; mj_2 = ``M"$\;
    $mn_1 = ``S"; mn_2 = ``M"$\;}
    
   \uElseIf{$fig=``Figure 3"$}{
    $mj_1 = ``M"; mj_2 = ``P"$\;
    $mn_1 = ``M"; mn_2 = ``S"$\;}

 \uElseIf{$fig=``Figure 4"$}{
    $mj_1 = ``P"; mj_2 = ``M"$\;
    $mn_1 = ``M"; mn_2 = ``S"$\;}
\end{algorithm}
\vspace{0.5cm}

$\bullet$ Subalgorithm Decision:\\
This algorithm determines major and minor sets for each prepositions (A, E, I and O) of major and minor premises by using $Set\_Interpretation()$ subalgorithm. We obtain premises conclusion via $Syllogistic\_Mapping()$ using major set and minor set values with respect to Table \ref{tab-16}. Later, for the analysed figure the premises conclusion set is compared to all conclusion sets under the corresponding state. If these are equal to each other, then the algorithm prints ``valid" output for the related syllogism.\\

\begin{algorithm}[H]
\DontPrintSemicolon
\caption{Decision\label{A3}}
\KwData{The set interpretations of major and minor premises of syllogisms.}
\KwResult{Obtain the Conclusion set of syllogisms and make a decision for syllogisms whether ``Valid" or ``Invalid".}
\BlankLine
\emph{\textbf{Decision(cond,fig)}} 

\ForEach{$mj\_prep$ in $Prepositions\{A, E, I, O\}$}{
	$major\_set = Set\_Interpretation(mj\_prep,``major",cond)$\;
	\ForEach{$mn\_prep$ in $Prepositions\{A, E, I, O\}$}{
		$minor\_set = Set\_Interpretation(mn\_prep,``minor",cond)$\;
		$premises\_conclusion=Syllogistic\_Mapping(major\_set,minor\_set)$\;
		\ForEach{$conc\_prep$ in $Prepositions\{A, E, I, O\}$}{
		\If {$premises\_conclusion=Conc[cond][conc\_prep]$}{$Print \  mj\_prep\ \& \ mn\_prep\ \& \ conc\_prep\ \& \ ``-Valid"$}
		}
	}
}
\end{algorithm}
\vspace{0.5cm}

$\bullet$ Subalgorithm Set\_Interpretation:\\
In this algorithm, temp set is determined for premise type and premise preposition as unconditional state. \\
- If the state is ``S exists" and the premise type is ``minor" then new temp set is determined by taking the transpose of temp set and it returns result of the subalgorithm $Syllogistic\_Mapping()$ which gets inputs as the constant set and the new temp set, respectively.\\
- If the state is ``M exists" and the premise type is ``minor" then it returns result of the subalgorithm $Syllogistic\_Mapping()$ which gets inputs as the temp set and the constant set, respectively.\\
- If the state is ``P exists" and the premise type is ``major" then new temp set is determined by taking the transpose of temp set and it returns result of the subalgorithm $Syllogistic\_Mapping()$ which gets inputs as the constant set and the new temp set, respectively.\\
- If the state is ``Unconditional" then it just returns the temp set.

\begin{algorithm}[H]
\DontPrintSemicolon
\caption{Set Interpretation\label{A4}}
\KwData{The specified premise preposition, premise type and state}
\KwResult{The conclusion set}
\BlankLine

\emph{\textbf{$Set\_Interpretation(premise\_prep, premise\_type, cond)$}} \;
Determine $Temp\_set$ using premise\_type and premise\_prep for Unconditional state with respect to the Diagram\;
\If{$premise\_type=``minor"$}{
    \If{$cond=``S \ Exists"$}
    {$NTemp\_set=transpose\ the\ diagram\ of\ Temp\_set$\;
    $return \  Syllogistic\_Mapping(Const\_set, NTemp\_set)$}
    \If{$cond=``M \ Exists"$} 
         {$return \  Syllogistic\_Mapping(Temp\_set, Const\_set)$}
}          
   
\If{$premise\_type=``major"$}{
    \If{$cond=``P \ Exists"$}
    {$NTemp\_set=transpose\ the\ diagram\ of\ Temp\_set$\;
    $return \  Syllogistic\_Mapping(Const\_set, NTemp\_set)$}
    }
$return \  Temp\_set$

\end{algorithm}

\section{Conclusion}

In this paper, we present a new effective algorithm for categorical syllogisms by using calculus system SLCD at the first time in literature. In accordance with this purpose, we explain categorical syllogisms by the help of Carroll's diagrams and we find unconditionally valid syllogisms and conditionally valid syllogisms via this algorithmic approach. As a result, our aim in this paper is to design an algorithm to contribute to researchers getting into the act in different areas of science used categorical syllogisms such as artificial intelligence, engineering, computer science and also mathematics.


\bibliography{mybibfile}

\end{document}